# On the Performance of Forecasting Models in the Presence of Input Uncertainty

Hossein Sangrody[1], *Student Member, IEEE,* Morteza Sarailoo[1], *Student Member, IEEE*, Ning Zhou[1], *Senior Member, IEEE*, Ahmad Shokrollahi [2], Elham Foruzan[3], *Student Member, IEEE*

[1] Electrical and Computer Engineering Department, State University of New York at Binghamton, Binghamton, NY 13902, USA { habdoll1, msarail1, ningzhou}@binghamton.edu
[2]Mazandaran Regional Electric Company, Sari, Iran, Ashokrollahi@mazrec.co.ir
[3] Electrical and Computer Engineering Department, University of Nebraska Lincoln, Lincoln, NE, USA 68503, elham.foruzan@huskers.unl.edu

*Abstract*—**Nowadays, with the unprecedented penetration of renewable distributed energy resources (DERs), the necessity of an efficient energy forecasting model is more demanding than before. Generally, forecasting models are trained using observed weather data while the trained models are applied for energy forecasting using forecasted weather data. In this study, the performance of several commonly used forecasting methods in the presence of weather predictors with uncertainty is assessed and compared. Accordingly, both observed and forecasted weather data are collected, then the influential predictors for solar PV generation forecasting model are selected using several measures. Using observed and forecasted weather data, an analysis on the uncertainty of weather variables is represented by MAE and bootstrapping. The energy forecasting model is trained using observed weather data, and finally, the performance of several commonly used forecasting methods in solar energy forecasting is simulated and compared for a real case study.**

## I. Introduction

Forecasting in both load and energy is an indispensable part of planning, operating, and controlling of a power system [1]. Nowadays, with the unprecedented penetration of renewable distributed energy resources (DERs), and their indispatchability, variability, and uncertainty, the necessity of efficient forecasting models is more vital than before [2-5]. Generally, when weather variables are considered in load or energy forecasting, the forecasting model is trained using observed weather data while in practice the forecasted weather data is applied in the forecasting model [6].

Many studies have been devoted to renewable DERs generation forecasting and improving their forecasting models to have more accurate results [7-8]. In [9] an intelligent method based on coding and image processing has been proposed to forecast wind speed and solar radiation. K-nearest neighbor algorithm has also been applied for solar photovoltaic (PV) generation forecasting in [10], where the prediction of weather and solar irradiance have been considered as predictors. Validation of different weather prediction models for several regions of the continental United State has been studies in [11] using SURFRAD ground measurement data. In [12], past power measurement, solar irradiance forecast, humidity, and temperature have been applied in an artificial neural network based method to forecast solar power. These studies are sample of many studies in DERs generation forecasting which have laid a solid ground in this field and at the same time reveals the need for considering the effect of weather forecasting uncertainties on generation forecasting results.

In this study, the performance of widely used forecasting methods on the uncertainty of the weather forecast is assessed. The solar PV panel installed on the rooftop of Engineering and Science buildings at the State University of New York at Binghamton is considered as a case study. Accordingly, the observed weather data along with the forecasted weather data for six successive days ahead are collected and an analysis on selecting influential predictors for energy forecasting is done on weather variables. When the influential predictors are selected, error analyses using common error metrics and bootstrapping are done. The energy forecasting models are trained using observed weather data. Finally, the performance of energy forecasting models using forecasted weather data is compared.

The rest of the paper is organized as follows. In Section II and III, the process of weather data acquisition, and analysis on error in weather data are elaborated in details, respectively. Section IV explains several forecasting methods applied in this paper. The quantification of error analysis using common error metrics and bootstrapping is discussed in Section V. Simulation results regarding all the discussions in the previous sections are represented in Section VI. At the end, the conclusions are drawn in Section VII.

## II. WEATHER DATA ACQUISITION

In solar PV generation forecasting, weather variables, time, and date are the most common inputs which are directly or indirectly applied in forecasting models. Among weather variables, sky cover, relative humidity, dew point, and temperature are the major predictors of forecasting model although other variables may also be considered in some studies [13]. To assess the effects of weather predictors' uncertainty in forecasting model, weather data for both forecasted and observed values are collected from National Weather Service of National Oceanic and Atmospheric Administration (NOAA), which is available for the public [14]. At this service, the forecasted and overserved data of weather variables are accessible for most of local areas in the US with an hourly resolution. The National Weather Service provides forecasting of aforementioned weather variables with hourly resolution for 6 consecutive days ahead. Both the observed and forecasted weather variables include wind, visibility, weather condition, sky condition, temperature, dew point, relative humidity, pressure, and precipitation; however, the sky cover is categorized in 5 categories in overserved data while in forecasted data, it is quantified by continuous numbers. Therefore, an adjustment on the observed value of sky cover is required to transform the classified values into numbers which makes it feasible to use both observed and forecasted value in the same forecasting model.

## III. SELECTING PREDICTORS OF SOLAR ENERGY FORECASTING MODEL

In this section, an analysis on selecting the influential predictors in solar energy forecasting model is conducted. To drive the influential predictors in forecasting model, there are several measures by which the performance of forecasting models and their predictors can be compared. With such measures, forecasting models corresponding to difference predictors are compared and the best predictors are selected for forecasting.

One of the simple method is to check *p*-values regarding to the predictors in the multi linear regression results and dismissing predictors with small *p*-value (less than 0.05). However, this method is not recommended since there might be correlation between predictors. Instead, there are six methods which are widely accepted in selecting predictors [15].

$\bar{R}^2$ (also called *Adjusted $R^2$*) is the adjusted form of $R^2$ is represented as follows.

$$\bar{R}^2 = 1 - (1 - R^2)\frac{N-1}{N-k-1} \quad (3)$$

Where $N$ is the number of observations, and $k$ is the number of predictors. Larger value of $\bar{R}^2$ for a model indicates better performance of its predictors in compared with other models.

*Akaike's Information Criterion* (*AIC*) is defined by (4).

$$AIC = N \log\left(\frac{SSE}{N}\right) + 2(k+2) \quad (4)$$

Where $SSE$ is the minimum sum of squared errors ($e_i$) corresponding to residuals obtained by comparing between forecasted and observed target variables. Note that although there might be different definition for $AIC$ in other literatures, the result should conclude the predictor set.

In contrast with $\bar{R}^2$, the minimum value of $AIC$ corresponds to best forecasting model and its predictors.

When the number of observations is too small, the $AIC$ results in too many predictors, so *Corrected Akaike's Information Criterion* (*AICc*) defined by (5) is applied for such a case [16].

$$AICc = AIC + 2k * \frac{k+1}{N-k-1} \quad (5)$$

*Bayesian Information Criterion* (*BIC*) is another measure for selecting predictors which is defined as follows.

$$BIC = N \log\left(\frac{SSE}{N}\right) + 2(k+2) \log N \quad (6)$$

Finally, *Cross Validation* (*CV*) is the other measure in this paper for selecting predictors formulated as follows.

$$CV = \frac{1}{N} \sum_{1}^{N} [e_i/(1-h_i)]^2 \quad (7)$$

Where $h_i$ is the diagonal value of *hat-matrix* (**H**). **H** is defined as (8) where **X** the matrix form of predictors which is elaborated in the next section.

$$H = X(X^T X)^{-1} X^T \quad (8)$$

In addition, *Stepwise regression* can be used for selecting predictors; however, although it always derives good predictors, there is no guarantee to gives the best possible set of predictors. In this paper, all five aforementioned methods are applied in selecting predictors.

## IV. FORECASTING METHODS

There are lots of methods in forecasting and it is not possible to mention a short description of them in this paper. However, this paper implements some of the most commonly used methods in forecasting. Forecasting methodologies and analysis can be classified into three main categories of time series analysis, machine learning, and hybrid methods. Popular methods in time series analysis category include multiple linear regression (MLR), quantile regression (QR), auto-regressive moving average (ARMA), autoregressive integrated moving average (ARIMA). Methods developed under machine learning category includes [17] artificial neural network (ANN), support vector regression (SVR), and fuzzy logic (FL). In hybrid method, a combination of several methods with heuristic algorithm is implemented in forecasting. Among the aforementioned methods, MLR, QR, ANN, and SVR are selected as the most commonly methods in forecasting to apply in this study. In addition, Ensemble (Ens) method in which several forecasting models are combined together is also considered as the hybrid method implemented in this paper. In the following section, a short description of the aforementioned methods is discussed [18].

### A. MLR Method

In the MLR method, the relationship between predictors and target is modeled by a linear equation. Accordingly, in each observation, the predictor ($x_i$) is associated to the target ($y_i$) with a constant coefficient ($\beta$). Such a description is shown in (9) where at observation $i$, the target $y_i$ is represented by linear fitting equation for $k$ predictors [19].

$$y_i = \beta_0 + x_{i1}\beta_1 + \cdots + x_{ik}\beta_k + e_i$$
$$for \quad i = 1, 2, \dots, N \quad (9)$$

Where $N$ is the number of observations and $e_i$ is the error.

Similarly, assuming predictors set as vector $X$ for $N$ number of observations and target vector of $Y$, equation (9) is represented as (10) where $\boldsymbol{\beta}$ and $\boldsymbol{e}$ are coefficient and error vectors, respectively.

$$Y = X^T.\boldsymbol{\beta} + \boldsymbol{e} \quad (10)$$

The least square (LS) method results in estimated coefficient vector ($\widehat{\boldsymbol{\beta}}$) in (11) by minimizing the summation of squared errors.

$$\widehat{\boldsymbol{\beta}} = \underset{\beta}{\text{argmin}} \left\{ \sum_{i=1}^{N} e_i^2 \right\} \quad (11)$$

### B. QR Method

In contrast with the MLR method in which there is no difference between errors for under- or over-fitting, the QR considers them differently and derives the regression coefficients by minimizing the sum of over and under fitting errors values with different penalties. In other words, different from MLR, QR considers directional errors instead of the squared errors and as shown in (12), it models the relationship between predictors and conditional percentile of target.

$$Q_\tau(Y|X) = X^T.\boldsymbol{\beta}_\tau \quad (12)$$

Therefore, as shown in (13), for the conditional quantile of $\tau$ ($0 < \tau < 1$), the estimated coefficients $\widehat{\boldsymbol{\beta}}_\tau$ is calculated as follows.

$$\widehat{\boldsymbol{\beta}}_\tau = \underset{\beta}{\text{argmin}} \left[ \sum_{i:\, y_i \geq X_i^T \boldsymbol{\beta}_\tau}^{N} \tau.|y_i - X_i^T \boldsymbol{\beta}_\tau| + \sum_{i:\, y_i < X_i^T \boldsymbol{\beta}_\tau}^{N} (1 - \tau).|y_i - X_i^T \boldsymbol{\beta}_\tau| \right] \quad (13)$$

### C. ANN Method

The ANN is an efficient method to model a simple linear or a complex nonlinear system. Like a black-box model, in an ANN model, there is no need to derive the closed-form equations of a system or to figure out the complex relationship between predictors and target variables. Such a simplicity of application in one hand and the efficiency of the ANN model in forecasting in the other hand make this method one of the most pervasive and popular methods in forecasting. An ANN model consists of three layers of input, output, and hidden layers. In each layer, there are several neurons which are connected to other neuron located in other layers by weighted connections. The number of neurons in input and output layers are respectively the same as the number of predictors and targets; however, the number of hidden layers and their neurons are specified by user. For many cases, one or two hidden layers provide fairly good results [20].

### D. SVR Method

The SVR is the regression version of support vector machine (SVM). As shown in (14), in SVR the forecasting model is trained by minimizing the sum of training error $\sum_{i=1}^{N}(\xi_i + \xi_i^*)$ and regulation term $\frac{1}{2}\|w\|^2$ subjected to constraints in (15).

$$\frac{1}{2}\|w\|^2 + C \sum_{i=1}^{N} (\xi_i + \xi_i^*) \quad (14)$$

$$Constraints: \begin{cases} y_i - (w^T \phi(x_i) + b) \leq \varepsilon + \xi_i \\ (w^T \phi(x_i) + b) - y_i \leq \varepsilon + \xi_i^* \end{cases} \quad \xi_i, \xi_i^* \geq 0 \quad (15)$$

Where $N$ is the number of observations, $\varepsilon$ is the margin of tolerance, $b$ is intercept, and $\xi_i$ and $\xi_i^*$ are upper and lower training errors associated to $\varepsilon$, respectively. In addition, $\phi$ is the kernel function which transforms $x_i$ to higher dimensional space.

### E. Ens method

An ensemble method is based on the assumption that although several forecasting methods may not result in satisfying forecasting results individually, they have their own benefits at some points and if they are combined efficiently, it may result in a better forecasting model. Accordingly, ensemble method combines several methods together to derive an efficient forecasting model. In ensemble method, while every implemented forecasting model provides its unique forecast, an ensemble model combines the results of all forecasting models together using a weighted average to achieve better accuracy. With proper weightings, the ensemble method bears a potential to obtain a more accurate forecast than its individual methods.

## V. QUANTIFICATION OF ERRORS AND UNCERTAINTY

Forecasting results for both weather variables and solar energy can be quantified with several metrics. The commonly used metrics are the mean absolute percentage error (MAPE), mean absolute error (MAE), mean squared error (MSE), and root-mean-square error (RMSE) [18]. Among the aforementioned metrics, MAPE is considered the most commonly used metric in load and energy generation forecasting which is represented by (16).

$$\text{MAPE} = \frac{1}{N} \sum_{i=0}^{N} \left| \frac{y_i - \hat{y}_i}{y_i} \right| \times 100 \quad (16)$$

Where $N$ is the total number of observations, $y_i$ is the real value, and $\hat{y}_i$ is the forecasted value. As shown, in this metric, the error is divided by the real value ($y_i$), therefore this metric is not applicable for the weather variables forecasting quantification which get zero value. Accordingly, MAE defined by (17) is applied to quantify error in weather variables.

$$\text{MAE} = \frac{1}{N} \sum_{i=0}^{N} |y_i - \hat{y}_i| \quad (17)$$

In addition, in statistical inference, bootstrapping is an efficient numerical method which is used to infer statistical parameters of population from a sample [21]. If the sample is drawn properly, bootstrapping is able to extract a fairly good approximation of the population's statistics like mean or standard deviation [22]. Bootstrapping is based on resampling of a sample by drawing and replacement of a subsample and it does not require any assumption about distribution of sample; however, sample should be selected properly from the population to be a good representative of the population and with sufficient size. In this study, the number of resampling in bootstrapping is considered 2500 times. [23]. Bootstrapping also provides the uncertainty of statistical parameters which is represented by confidence interval (CI). With such a CI, it claims that the true value of the statistical parameter of population is located within the CI with a predefined probability. Such a probability is usually quantified with 95% which means that the true statistical parameter of the population is assured to be located within the CI with the probability of 0.95 [21]. Accordingly, mean and standard deviation of the population are estimated with a value along with the CI.

## VI. SIMULATION RESULTS

In this case study, for the simulation of predictors selection and forecasting methods, MATLAB® software is used. The case study for the analysis on weather variable and solar energy is the State University of New York at Binghamton where the solar PV generation data on the rooftop of the Engineering and Science buildings during May 2016 to October 2016 are applied for this case. In addition, the weather data for both observed and forecasted values are collected from the National Weather Service of National Oceanic and Atmospheric Administration (NOAA) for the case study and at the same studying time[1]. Both data for the solar PV generation and weather variables have hourly resolution. The weather forecast provided by the National Weather Service for six days ahead. Accordingly, for a value of solar energy data in a specific hour of during daytime, there are one observed weather variables set and six forecasted data sets corresponding to that hour during six days ahead.

As mentioned, the weather data provided by the National Weather Service include serval weather variables both for the observed and forecasted variables. However, among all the variables common between overserved and forecasted groups, the most potential predictors for the case of solar energy forecasting are sky cover, dew point, relative humidity, and temperature. However, the sky cover in the observed data set is represented in categories as shown in the first column of Table I, whereas in forecasted data it is provided by continuous percent values. The categories of the sky cover are shown in the second column which is also represented by percentage in the third column. Accordingly, to apply the observed data for the sky cover in forecasting model, the categorized observed value of the sky cover is adjusted by the percent categories as shown in the fourth column of the Table I [14].

Before applying four weather variables in energy forecasting models, the discussions of selecting predictors are simulated using five measures mentioned in Section III. The result of simulation is shown in Table II wherein the first column, 15 states of selecting predictors (sky cover (S), dew point (D), relative humidity (R), and temperature (T)) are shown with their abbreviation letters. In the other columns, the values of selecting predictors measures are shown along with

---

[1] The data is available upon request to the first author.

---

RMSE. Similar to MAPE and MAE, RMSE is a measure to check error in forecasting which is calculated by (18).

$$\text{RMSE} = \sqrt{\frac{1}{N}\sum_{i=0}^{N}(y_i - \hat{y}_i)^2} \qquad (18)$$

TABLE I. OBSERVED SKY COVER CATEGORIES

| Sky Condition | Opaque Cloud Coverage | Opaque Cloud Coverage (%) | Percentage Category (%) |
|---|---|---|---|
| Clear | 1/8 and less | Sky Cover < 12.5 | 0 |
| Mostly Clear | 1/8 to 3/8 | 12.5 ≤ Sky Cover < 37.5 | 25 |
| Partly Cloudy | 3/8 to 5/8 | 37.5 ≤ Sky Cover < 62.5 | 50 |
| Mostly Cloudy | 5/8 to 7/8 | 62.5 ≤ Sky Cover < 87.5 | 75 |
| Cloudy | 7/8 to 8/8 | 87.5 ≤ Sky Cover | 100 |

As shown in this table, most of methods confirm using all the predictors set (SDRT), so the SDRT predictors are selected in energy forecasting models. In Table II, in some measures, predictor set of DRT is preferred on SDRT with a minor difference. The reason for such a difference is the fact that the observed value for sky cover is originally represented categories and it has been adjusted in Section VI. Such an adjustment on categorized observed sky cover variable causes errors in its numerical percent value for comparing with forecasted value.

TABLE II. SELECTING PREDICTORS

|  | RMSE | $\bar{R}^2$ | AIC | AICc | BIC | CV |
|---|---|---|---|---|---|---|
| SDRT | 12.30 | 0.67 | 1167.71 | 1168.13 | 1182.70 | 221.96 |
| DRT | 12.33 | 0.67 | 1167.52 | 1167.80 | 1179.51 | 221.98 |
| SDR | 13.01 | 0.63 | 1183.52 | 1183.80 | 1195.50 | 608.00 |
| DR | 13.06 | 0.62 | 1183.56 | 1183.73 | 1192.55 | 622.31 |
| SRT | 13.26 | 0.61 | 1189.11 | 1189.39 | 1201.10 | 278.23 |
| RT | 13.30 | 0.61 | 1189.05 | 1189.22 | 1198.04 | 278.65 |
| SR | 13.66 | 0.59 | 1196.75 | 1196.92 | 1205.75 | 1774.83 |
| D | 13.68 | 0.59 | 1196.21 | 1196.29 | 1202.21 | 1864.89 |
| SDT | 14.30 | 0.55 | 1211.45 | 1211.73 | 1223.43 | 223.64 |
| DT | 14.46 | 0.54 | 1213.62 | 1213.79 | 1222.61 | 228.55 |
| ST | 17.53 | 0.32 | 1270.61 | 1270.78 | 1279.60 | 319.18 |
| SD | 18.83 | 0.22 | 1291.83 | 1292.00 | 1300.83 | 736.36 |
| S | 18.95 | 0.21 | 1292.72 | 1292.80 | 1298.71 | 4609.27 |
| T | 19.55 | 0.16 | 1301.98 | 1302.06 | 1307.97 | 385.69 |
| R | 20.61 | 0.07 | 1317.57 | 1317.65 | 1323.57 | 846.78 |

As mentioned, the weather forecasted variables are provided by the National Weather Service of NOAA for six days ahead. Considering the observed values of the weather variables as real values, and historical forecasted values of weather variables for six days ahead as forecasted values, the MAEs corresponding to each day ahead is calculated using (17). Table III represents the results of the MAE statistics i.e. mean and standard deviation (Std) in forecasting day (D #no represents the number of day) for sky cover, dew point, relative humidity, and temperature. In addition, the result of *Bias* defined by (19) is also shown in the table.

$$Bias = \frac{1}{N}\sum_{i=0}^{N}(y_i - \hat{y}_i) \quad (19)$$

Where, $N$ is the number of observations, $y_i$ is the observed weather variable, and $\hat{y}$ is the forecasted weather variable.

The statistics represented in the Table III are closely estimated to true value of the population using bootstrapping. As seen, the error is growing as the forecasting span is larger. As an example, the result of bootstrapping on the mean of MAEs for the sky cover is illustrated in Fig. 1 which clearly shows the error is rising in long term. In addition, the sign of *Bias* indicates the direction of the error. In other words, the positive sign indicates underestimating whereas the negative represent overestimating in forecasting. Accordingly, there is an overestimating in weather forecasting provided by the NOAA.

TABLE III. STATISTICS OF WEATHER FORECAST ERROR

| Type | Statistics | D #1 | D #2 | D #3 | D #4 | D #5 | D #6 |
|---|---|---|---|---|---|---|---|
| S | Bias | -11.1 | -8.5 | -8.4 | -7.5 | -7.1 | -8.4 |
| S | Mean | 26.52 | 27.49 | 29.80 | 32.23 | 34.28 | 35.64 |
| S | Std | 0.59 | 0.57 | 0.53 | 0.5 | 0.49 | 0.48 |
| D | Bias | -1.93 | -2.18 | -2.44 | -2.33 | -2.16 | -2.24 |
| D | Mean | 2.78 | 3.21 | 3.61 | 3.81 | 4.12 | 4.54 |
| D | Std | 0.06 | 0.07 | 0.8 | 0.8 | 0.85 | 0.09 |
| R | Bias | 0.48 | -0.32 | -0.18 | -0.19 | 0.14 | -0.19 |
| R | Mean | 8.35 | 9.32 | 10.30 | 10.30 | 11.08 | 11.56 |
| R | Std | 0.07 | 0.07 | 0.07 | 0.07 | 0.08 | 0.08 |
| T | Bias | -2.0 | -1.7 | -1.8 | -1.9 | -1.7 | -1.7 |
| T | Mean | 3.34 | 3.36 | 3.63 | 3.67 | 3.87 | 4.07 |
| T | Std | 0.18 | 0.19 | 0.23 | 0.22 | 0.24 | 0.23 |

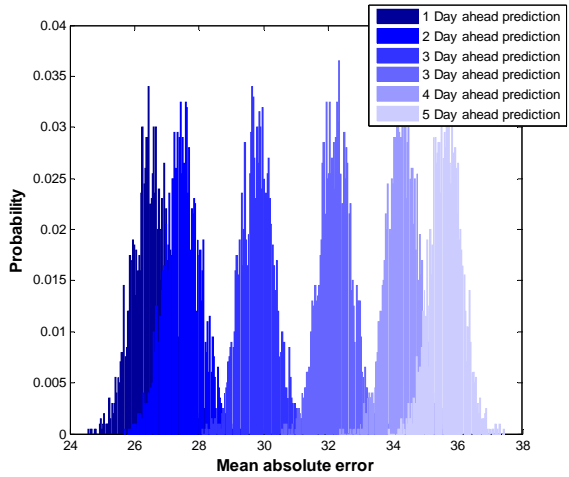

Figure 1. Error in the forecasted sky cover

In Table IV, the result of 95% CI for mean value of MAE which is derived by bootstrapping is depicted. Here in this table, each cell has two values which represent lower and upper bounds of 95% CI, respectively.

TABLE IV. 95% CONFIDENCE INTEVAL IN MEAN OF MAE

| Type | D #1 | D #2 | D #3 | D #4 | D #5 | D #6 |
|---|---|---|---|---|---|---|
| S | 25, 28 | 26, 29 | 28, 31 | 31, 33 | 33, 36 | 34, 37 |
| D | 2.6, 3 | 3, 3.4 | 3.4, 4 | 3.8, 4.1 | 3.9, 4.3 | 4.2, 5 |
| R | 8.1, 8.7 | 9.1, 9.5 | 10, 10.5 | 10, 10.5 | 10.8, 11 | 11, 12 |
| T | 2.8, 3.9 | 2.8, 3.9 | 3, 4.3 | 3, 4.3 | 3.1, 4.6 | 3.4, 4.8 |

Note that the error regarding the forecasted weather variable is obtained by comparing the historical forecasted weather and observed values; however, the paper is followed by the assumption that a forecaster is not sure about the uncertainty in forecasted weather variables and this study aims to analyze the robustness of the commonly used forecasting methods.

As mentioned, for the case study, the solar energy data is collected from the solar PV panels installed on the rooftop of the Engineering and Science buildings at the State University of New York, at Binghamton. Fig. 2 illustrates the solar energy generation for the case study during the studying period.

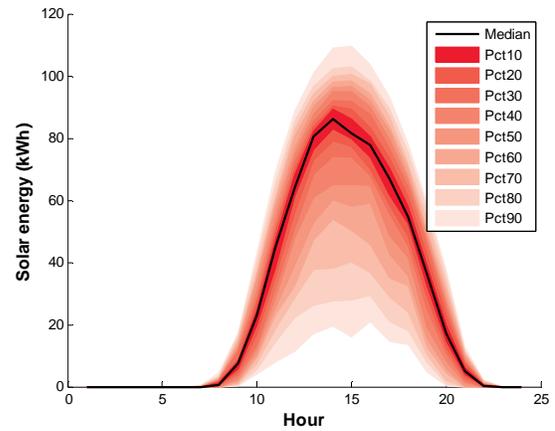

Figure 2. Solar energy generation for the case study

Here, in this case, the energy data has logged in hourly resolution; however, for the energy forecasting model analysis, only peak value of solar energy in each day is selected as target values. Consequently, the weather variables corresponding to the peak time are also extracted for predictors. By selecting predictor set of SDRT, the forecasting models using MLR, QR, ANN, SVR, and Ens methods are applied for the daily peak energy generation forecasting. In this study, the quantile regression is set at 50[th] quantile and the ANN is a feed forward supervised learning model, with the Levenberg-Marquardt algorithm for training, and one hidden layer with ten neurons. In addition, LIBSVM library is used for SVR method [24].

Using aforementioned methods, the forecasting model is trained using observed value of weather variables as predictors and solar energy generation as target. Then, the forecasted weather variable data for six successive days are applied in the trained models and the results are analyzed by MAPEs. The result of the aforementioned process is illustrated in Fig. 3. As shown, errors in all forecasting methods are rising in long term. This result is not surprising because of higher errors in the predictors in longer term. However, the MLR has the best forecasting in compared other methods with the presence of uncertainty in predictors although for the observed weather predictors the performance of this method is not significant. In addition, as illustrated, Ens, and ANN have also fairly satisfying results in the presence of error in the predictors.

Note that, the error in forecasting with the observed weather data is partially because of the fact that the observed sky cover variable is not continuous numbers like forecasted sky cover and it already adjusted from five categories to five numbers which imposes large uncertainties to this predictor.

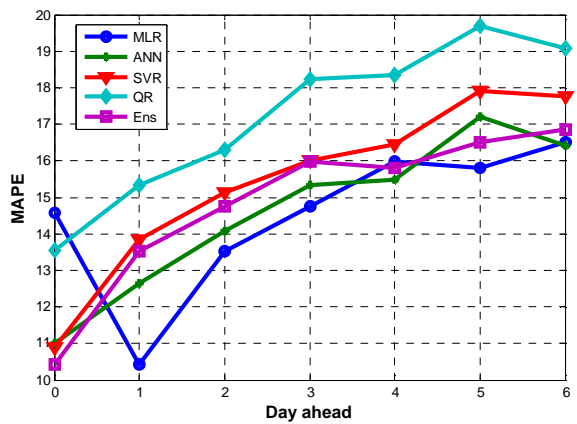

Figure 3. Results of MAPE for different forecasting methods

## VII. CONCLUSION

In this paper, the performance of widely used forecasting methods in the presence of uncertainty in weather data is studied. The most common weather variables including sky cover, dew point, relative humidity, and temperature are considered as candidate predictors for the solar PV generation forecasting. Then using the generation data of a solar PV panel as a case study, several predictive measures are applied to select the most influential predictors. The results of simulation indicate that all aforementioned weather variables are applied as predictors in energy forecasting model. An error analysis on the observed weather variable and forecasted value is done using MAE, Bias metrics, and bootstrapping. The result of error analysis concludes overestimating in weather forecasting data provided by the NOAA. For solar energy forecasting, the forecasting model is trained using the observed weather variables. Finally, the performance of the forecasting methods is simulated using both observed and forecasted data and the results are compared with MAPE. The result of forecasting indicates the high performance of MLR in the presence of uncertainty in weather forecasting data.